\documentclass[11pt,a4paper]{article}
\usepackage{naaclhlt2019}
\usepackage{times}
\usepackage{latexsym}
\usepackage{url}
\usepackage{amsmath}
\usepackage{amssymb}
\usepackage{tikz}
\usepackage{multirow}
\usepackage{adjustbox}
\usepackage{tabularx}
\usepackage{blindtext}
\usepackage{algorithm}
\usepackage[noend]{algorithmic}
\usepackage[font=small,skip=2pt]{caption}
\usetikzlibrary{positioning}
\usepackage{enumitem}
\usepackage[normalem]{ulem}
\usepackage{siunitx,tabularx,booktabs}

\aclfinalcopy 

\title{Evaluating Text GANs as Language Models}
\author{Guy Tevet\thanks{\;\;\; The authors contributed equally}$^{*,1,2}$ ~~~~~ Gavriel Habib$^{*,1,2}$ ~~~~~
Vered Shwartz$^3$ ~~~~~
Jonathan Berant$^{1,4}$ \\
\mbox{}\\
$^1$School of Computer Science, Tel-Aviv University \\
$^2$Department of Electrical Engineering, Tel-Aviv University \\
$^3$Computer Science Department, Bar-Ilan University\\
$^4$Allen Institute for Artificial Intelligence \\
\small{\texttt{\{guytevet@mail,gavrielhabib@mail,joberant@cs\}.tau.ac.il},~\texttt{vered1986@gmail.com}}}

\begin{document}
\maketitle

\begin{abstract}

Generative Adversarial Networks (GANs) are a  promising approach for text generation that, unlike traditional language models (LM), does not suffer from the problem of ``exposure bias". 
However, A major hurdle for understanding the potential of GANs for text generation is the lack of a clear evaluation metric. In this work, we propose to approximate the distribution of text generated by a GAN, which permits evaluating them with traditional probability-based LM metrics. 
We apply our approximation procedure on several GAN-based models and show that they currently perform substantially worse than state-of-the-art LMs. 
Our evaluation procedure promotes better understanding of the relation between GANs and LMs, and can accelerate progress in GAN-based text generation.
\end{abstract}

\section{Introduction}
\label{sec:intro}

Neural networks have revolutionized the field of text generation, in machine translation \cite{sutskever2014sequence,neubig2017neural,luong2015multi,chen2018best}, summarization \cite{see2017get}, image captioning \cite{you2016image} and many other applications \cite{goldberg2017neural}.

Traditionally, text generation models are trained by going over a gold sequence of symbols (characters or words) from left-to-right, and maximizing the probability of the next symbol given the history, namely, a language modeling (LM) objective. A commonly discussed drawback of such LM-based text generation is \emph{exposure bias} \cite{ranzato2015sequence}: during training, the model predicts the next token conditioned on the ground truth history, while at test time prediction is based on predicted tokens, causing a train-test mismatch. Models trained in this manner often struggle to overcome previous prediction errors.

Generative Adversarial Networks \cite{goodfellow2014generative} offer a solution for \emph{exposure bias}. Originally introduced for images, GANs leverage a discriminator, which is trained to discriminate between real images and generated images via an adversarial loss. In such a framework, the generator is not directly exposed to the ground truth data, but instead learns to imitate it using global feedback from the discriminator. This has led to several attempts to use GANs for text generation, with a generator using either a recurrent neural network (RNN) \cite{yu2017seqgan,guo2017long,press2017language,rajeswar2017adversarial}, or a Convolutional Neural Network (CNN) \cite{gulrajani2017improved,rajeswar2017adversarial}.

However, evaluating GANs is more difficult than evaluating LMs.
While in language modeling, evaluation is based on the log-probability of a model on held-out text, this cannot be straightforwardly extended to GAN-based text generation, because the generator outputs discrete tokens, rather than a probability distribution. Currently, there is no single evaluation metric for GAN-based text generation, and existing metrics that are based on n-gram overlap are known to lack robustness and have low correlation with semantic coherence \cite{semeniuta2018accurate}.

In this paper, we propose a method for evaluating GANs with standard probability-based evaluation metrics.
We show that the expected prediction of a GAN generator can be viewed as a LM, and suggest a simple Monte-Carlo method for approximating it. The approximated probability distribution can then be evaluated with standard LM metrics such as perplexity or Bits Per Character (BPC).

To empirically establish our claim, we implement our evaluation on several RNN-based GANs: \cite{press2017language,yu2017seqgan,guo2017long}. We find that all models have substantially lower BPC compared to state-of-the-art LMs. 
By directly comparing to LMs, we put in perspective the current performance of RNN-based GANs for text generation. 
Our results are also in line with  recent concurrent work by \newcite{caccia2018language}, who reached a similar conclusion by comparing the performance of textual GANs to that of LMs using metrics suggested for GAN evaluation.

Our code is available at: \url{http://github.com/GuyTevet/SeqGAN-eval} and  \url{http://github.com/GuyTevet/rnn-gan-eval}.

\section{Background}
\label{sec:related_work}

Following the success of GANs in image generation, several works applied the same idea to texts using convolutional neural networks \cite{gulrajani2017improved,rajeswar2017adversarial}, and later using RNNs \cite{press2017language,yu2017seqgan}. RNNs enable generating variable-length sequences, conditioning each token on the tokens generated in previous time steps. We leverage this characteristic in our approximation model (\S\ref{sec:lm_approximation}).

A main challenge in applying GANs for text is that generating discrete symbols is a non-differentiable operation. One solution is to perform a continuous relaxation of the GAN output, which leads to generators that emit a nearly discrete continuous distribution \cite{press2017language}. 
This keeps the model differentiable and enables end-to-end training through the discriminator. Alternatively, SeqGAN \cite{yu2017seqgan} and LeakGAN \cite{guo2017long} used policy gradient methods to overcome the differentiablity requirement. We apply our approximation to both model types.

\section{Evaluating GANs and LMs}
\label{sec:evaluation_methods}

\paragraph{LM Evaluation.} Text generation from LMs is commonly evaluated using probabilistic metrics. Specifically, given a test sequence of symbols $(t_1, \dots, t_n)$, and a LM $q$, the average cross-entropy over the entire test set is computed: $ACE = -\frac{1}{n} \sum_{i=1}^n{{\log_2 q(t_{i} \mid t_1 ... t_{i-1})}}$. For word-based models, the standard metric is perplexity: $PP = 2^{ACE}$, while for character-based models it is $BPC = ACE$ directly.

Intrinsic improvement in perplexity does not guarantee an improvement in an extrinsic downstream task that uses a language model. However, perplexity often correlates with extrinsic measures \cite{jurafsky2018speech}, 
and is the de-facto metric for evaluating the quality of language models today.

\paragraph{GAN-based Text Generation Evaluation.}
By definition, a text GAN outputs a discrete sequence of symbols rather than a probability distribution. As a result, LM metrics cannot be applied to evaluate the generated text. Consequently, other metrics have been proposed:

\begin{itemize}[noitemsep,parsep=0pt,partopsep=0pt,leftmargin=*]
    \item N-gram overlap: \cite{yu2017seqgan,press2017language}:
    Inspired by BLEU \cite{papineni2002bleu}, this measures whether n-grams generated by the model appear in a held-out corpus. A major drawback is that this metric favors conservative models that always generate very common text (e.g., \emph{``it is"}). To mitigate this, self-BLEU has been proposed \cite{lu2018neural} as an additional metric, where overlap is measured between two independently sampled texts from the model.
    
    \item LM score: The probability of generated text according to a pre-trained LM. This has the same problem of favoring conservative models.
    
    \item \newcite{zhao2017adversarially} suggested an indirect score by training a LM on GAN-generated text, and evaluating it using perplexity. The drawback in this setting is the coupling of the performance of the GAN with that of the proxy LM.
    
    \item \newcite{heusel2017gans} used Frechet InferSent Distance (FID) to compute the distance between distributions of features extracted from real and generated samples. However, this approach relies on a problematic assumption that features are normally distributed.
    
    \item \newcite{rajeswar2017adversarial} used a context-free grammar (CFG) to generate a reference corpus, and evaluated the model by the likelihood the CFG assigns to generated samples. However, simple CFGs do not fully capture the complexity of natural language.
    
    \item To overcome the drawbacks of each individual method, \newcite{semeniuta2018accurate} proposed a unified measure based on multiple evaluation metrics (N-grams, BLEU variations, FID, LM score variations and human evaluation). Specifically, they argue that the different measures capture different desired properties of LMs, e.g., quality vs. diversity.
    \item Following \newcite{semeniuta2018accurate}, and in parallel to this work, \newcite{caccia2018language} proposed a temperature sweep method that trades-off quality for diversity using a single parameter. Similar to our findings, they concluded that GANs perform worse than LMs on this metric.
    
\end{itemize}

Overall, current evaluation methods cannot fully capture the performance of GAN-based text generation models. While reporting various scores as proposed by \newcite{semeniuta2018accurate} is possible, it is preferable to have a single measure of progress when comparing different text generation models.

\section{Proposed Method}
\label{sec:proposed_method}

We propose a method for approximating a distribution over tokens from a GAN, and then evaluate the model with standard LM metrics. We will describe our approach given an RNN-based LM, which is the most commonly-used architecture, but the approximation can be applied to other auto-regressive models \cite{vaswani2017attention}.

\subsection{Language Model Approximation}
\label{sec:lm_approximation}

\begin {figure}[!t]
\centering
\resizebox {\columnwidth} {!} { %
\centering
\begin{tikzpicture}

\node[draw, outer sep=2, circle, minimum size=1cm] (x_t) {$x_t$};
\node[draw, outer sep=2, circle, minimum size=1cm] [right=of x_t](x_tp) {$x_{t+1}$};
\node[draw, outer sep=2, circle, minimum size=1cm] [right=of x_tp](x_tpp) {$x_{t+2}$};
\node[align=center, outer sep=2] [right=of x_tpp](x_tppp) {$x_{t+3}$};
\node[draw, align=center, outer sep=2, minimum size=0.8cm] (unit_t) [above=of x_t] {};
\node[draw, align=center, outer sep=2, minimum size=0.8cm] (unit_tp) [above=of x_tp] {};
\node[draw, align=center, outer sep=2, minimum size=0.8cm] (unit_tpp) [above=of x_tpp] {};
\node[align=center, outer sep=2] (h_t) [left=of unit_t] {$h_{t}$};
\node[align=center, outer sep=2] (h_tppp) [right=of unit_tpp] {$h_{t+3}$};
\node[draw, outer sep=2, circle, minimum size=1cm] (y_t) [above=of unit_t] {$o_t$};
\node[align=center, outer sep=2] [left=of y_t](y_tm) {$o_{t-1}$};
\node[draw, outer sep=2, circle, minimum size=1cm]  (y_tp) [right=of y_t] {$o_{t+1}$};
\node[draw, outer sep=2, circle, minimum size=1cm]  (y_tpp) [right=of y_tp] {$o_{t+2}$};

\path[->] (x_t) edge (unit_t);
\path[->] (unit_t) edge (y_t);
\path[->] (x_tp) edge (unit_tp);
\path[->] (x_tpp) edge (unit_tpp);
\path[->] (unit_tp) edge (y_tp);
\path[->] (unit_tpp) edge (y_tpp);
\path[->] (h_t) edge (unit_t);
\path[->] (unit_t) edge (unit_tp);
\path[->] (unit_tp) edge (unit_tpp);
\path[->] (unit_tpp) edge (h_tppp);

\path[->,dashed] (y_tm) edge (x_t);
\path[->,dashed] (y_t) edge (x_tp);
\path[->,dashed] (y_tp) edge (x_tpp);
\path[->,dashed] (y_tpp) edge (x_tppp);

\end{tikzpicture} 
}
\caption{Generator recurrent connections. $\{h_t\}$ is the internal state sequence and $\{o_t\}$ is the generator prediction sequence (one-hot). During inference, the outputs $\{o_t\}$ are fed back as the input for the next time step (dashed lines). During LM approximation, the input $\{x_t\}$ is a sequence of one-hot vectors from the test set.}
\label{fig:infernce}
\end {figure}
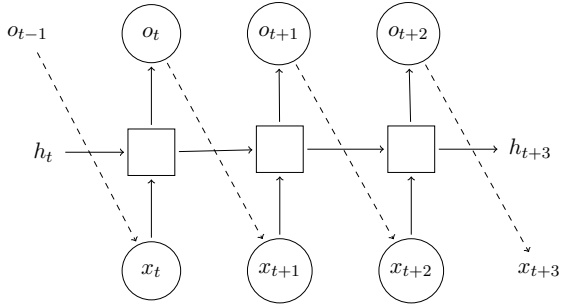

The inputs to an RNN at time step $t$, are the state vector $h_t$ and the current input token $x_t$. The output token (one-hot) is denoted by $o_t$. In RNN-based GANs, the previous output token is used at inference time as the input $x_t$ \cite{yu2017seqgan,guo2017long,press2017language,rajeswar2017adversarial}. In contrast, when evaluating with BPC or perplexity, the gold token $x_t$ is given as input. Hence, LM-based evaluation neutralizes the problem of exposure bias addressed by GANs. Nevertheless, this allows us to compare the quality of text produced by GANs and LMs on an equal footing. Figure~\ref{fig:infernce} illustrates the difference between inference time and  during LM approximation.

We can therefore define the generator function at time step $t$ as a function of the initial state $h_0$ and the past generated tokens $(x_0 \dots x_t)$, which we denote as $o_t=G_t(h_0,x_0...x_t)$ ($x_0$ is a start token). Given a past sequence $(x_0 \dots x_t)$, $G_t$ is a stochastic function: the stochasticity of $G_t$ can be gained either by using a noise vector as the initial state $h_0$ \cite{press2017language}, or by sampling from the GAN's internal distribution over possible output tokens  \cite{yu2017seqgan,guo2017long}.
Since $h_0$ is constant or a noise vector that makes $G_t$ stochastic, we can omit it to get $G_t(x_0 \dots x_t)$. In such a setup, the expected value $\mathbb{E}[G_t(x_0 \dots x_t)]$ is a distribution $q$ over the next vocabulary token $a_t$:
\begin{align*}
\vspace*{-3pt}
q(a_t \mid a_0 \dots a_{t-1}) &= \{\mathbb{E}[G_t(x_0 \dots x_t)]\}_{a_t}  
\vspace*{-3pt}
\end{align*}

\vspace*{-3pt}
To empirically approximate $q$,  we can sample from it $N$ i.i.d samples, and compute an approximation $\tilde{G}_{t,N}=\frac{1}{N} \Sigma_{n=1}^N{g_{t,n}}$, where $g_{t,n}$ is one sample from $G_t(x_0...x_t)$. Then, according to the strong law of large numbers:
\begin{align}
\vspace*{-3pt}
\mathbb{E}[G_t(x_0 \dots x_t)] &= \lim_{N\rightarrow \infty} \tilde{G}_{t,N}
\label{eq:converge}
\vspace*{-3pt}
\end{align}
\vspace*{-3pt}

Given this approximate LM distribution, we can evaluate a GAN using perplexity or BPC. We summarize the evaluation procedure in Algorithm~\ref{algorithm}.\footnote{Our evaluation algorithm is linear in the length of the test set and in the number of samples $N$.}

\begin{algorithm}[t]
\caption{LM Evaluation of RNN-based GANs}
\label{algorithm}
\begin{algorithmic}[1]
{\footnotesize
\REQUIRE $G_t(\cdot)$: the generator function at time step $t$ \\ $(x_0, ..., x_t)$: previous gold tokens \\ $x_{t+1}$: the gold next token (as ground truth) \\ $f(\cdot, \cdot)$: a LM evaluation metric \\
$N$: number of samples
    \FOR{$n \gets 1$ to $N$}   
        \STATE{$g_{t,n} \longleftarrow$ sample from $G_t(x_0...x_t)$}
    \ENDFOR
    \STATE{$\tilde{G}_{t,N} = \frac{1}{N} \Sigma_{n=1}^N{g_{t,n}}$}
    \RETURN{$f(\tilde{G}_{t,N}, x_{t+1})$}
}
\end{algorithmic}
\end{algorithm}

\subsection{Approximation Bound}
\label{sec:approximation_bounds}

We provide a theoretical bound for choosing a number of samples $N$ that results in a good approximation of $\tilde{G}_{t,N}$ to $\mathbb{E}[G_t]$.

Perplexity and BPC rely on the log-probability of the ground truth token. Since the ground truth token is unknown, we conservatively define the bad event $B$ in which there exists $v \in V$ such that $|\{E[G_t]\}_v - \{\tilde{G}_{t,N}\}_v| > \gamma$, where $V$ is the vocabulary. We can then bound the probability of $B$ by some $\epsilon$. We define the following notations: 

\begin{enumerate}[noitemsep,parsep=0pt,partopsep=0pt,leftmargin=*]
    \item The probability of a token $a_t$ to be $v$ is $p_v \overset{\Delta}{=} q(a_t=v|a_0 \dots a_{t-1})=\{\mathbb{E}[G_t(x_0 \dots x_t)]\}_v$. 
    \item $\chi_{v,n} \overset{\Delta}{=}  \left\{g_{t,n}\right\}_{v}$ is a random variable representing the binary value of the $v$'th index of $g_{t,n}$ which is a single sample of $G_t$. Note that the average of $\chi_{v,n}$ over $N$ samples is $ X_{v} \overset{\Delta}{=} \frac{1}{N} \sum_{n=1}^N{\chi_{v,n}} = \left\{ \frac{1}{N} \sum_{n=1}^N{g_{t,n}}\right\}_{v} = \{\tilde{G}_{t,N}\}_{v}$.
\end{enumerate}

Using the above notation, we can re-define the probability of the bad event $B$ with respect to the individual coordinates in the vectors:

\begin{equation}
{\footnotesize
\begin{aligned}
Pr(B) &= Pr\left(\| \mathbb{E}[G_t] - \tilde{G}_{t,N} \|_{\infty}>\gamma\right) \\ &= Pr\left(\bigcup_{v \in V}|p_{v}-X_{v}| > \gamma\right) \overset{!}{<} \epsilon
\end{aligned}
}%
\end{equation}

We note that $\chi_{v,n} \sim Bernoulli(p_{v})$, and given that $\{\chi_{v,n}\}_{n=1}^N$ are i.i.d., we can apply the Chernoff-Hoeffding theorem \cite{chernoff1952measure,hoeffding1963probability}. According to the theorem, for every $v\in V$, $Pr(| X_{v}-p_{v}| > \gamma) < 2e^{-2N\gamma^2}$.
Taking the union bound over $V$ implies: 

\begin{equation}
\resizebox{0.88\columnwidth}{!}{%
        $Pr(B) = Pr \left(\bigcup_{v \in V}| X_{v}-p_{v}| > \gamma\right) < 2|V|e^{-2N\gamma^2} < \epsilon$%
        }
\end{equation}

Hence, we get a lower bound on $N$:

\begin{equation}
N > \frac{1}{2\gamma^2} \ln\left(\frac{2|V|}{\epsilon}\right)
\end{equation}

As a numerical example, choosing $\gamma=10^{-3}$ and $\epsilon=10^{-2}$, for a character-based LM over the text8 dataset, with $|V|=27$, we get the bound: $N>4.3\cdot10^6$.
With the same $\gamma$ and $\epsilon$, a typical word-based LM with vocabulary size $|V|=50,000$ would require $N>8.1\cdot10^6$.

In practice, probability vectors of LMs tend to be sparse \cite{kim2016character}. Thus, we argue that we can use a much smaller $N$ for a good approximation $\tilde{G}_{t,N}$. Since the sparsity of LMs is difficult to bound, as it differs between models, we suggest an empirical method for choosing $N$.

The approximation $\tilde{G}_{t,N}$ is a converging sequence, particularly over $\|\cdot\|_{\infty}$ (see Equation~\ref{eq:converge}). Hence, we can empirically choose an $N$ which satisfies $
\|\tilde{G}_{t,N-\alpha} - \tilde{G}_{t,N}\|_\infty < \gamma' , \;\;\;\; \alpha \in \mathbb{N}$. 
In Section~\ref{sec:eval} we empirically measure $\|\tilde{G}_{t,N-\alpha} - \tilde{G}_{t,N}\|_\infty$ as a function of $N$ to choose $N$. We choose a global $N$ for a model, rather than for every $t$, by averaging over a subset of the evaluation set.

\section{Evaluation}
\label{sec:eval}

\begin{table*}[!t]
\small
\centering
\begin{tabular}{ c c c c } 
\toprule
\textbf{Approach} & \textbf{Model} & \textbf{BPC} & \textbf{Approx. BPC} \\
\midrule
\multirow{5}{*}{\textbf{Language Models}} & mLSTM + dynamic eval \cite{krause2017dynamic} & 1.19 & \\ 
& Large mLSTM +emb +WN +VD \cite{krause2016multiplicative} & 1.27 & \\ 
& Large RHN \cite{zilly2016recurrent} & 1.27 & \\ 
& LayerNorm HM-LSTM \cite{chung2016hierarchical} & 1.29 & \\ 
& BN LSTM \cite{cooijmans2016recurrent} & 1.36 & \\ 
& Unregularised mLSTM \cite{krause2016multiplicative} & 1.40 & \\
\midrule
& SeqGAN - pre-trained LM \cite{yu2017seqgan} & 1.85 & 1.95\\
\midrule
\multirow{2}{*}{\textbf{GANs (LM Approximation)}} & SeqGAN - full adversarial training \cite{yu2017seqgan} & 1.99  & 2.08 \\ & Recurrent
GAN without pre-training \cite{press2017language} & & 3.31 \\ 
\midrule
\multirow{1}{*}{} & Uniform Distribution & 4.75 & \\
\bottomrule
\end{tabular}
\caption{Test set evaluation of different character-based models on the text8 dataset. State-of-the-art results are taken from \url{https://github.com/sebastianruder/NLP-progress/blob/master/language_modeling.md}. The uniform distribution is equivalent to guessing the next character out of $|V| = 27$ characters.}
\label{tab:results}
\end{table*}

\begin{table*}[!t]
\hspace*{-12pt}
\centering
\small
\begin{tabular}{p{1.6cm}@{\hskip 0.05in}p{14.5cm}}
\toprule
Model & Samples \\
\midrule
\multirow{3}{1.6cm}{SeqGAN\\{\scriptsize Pre-trained LM}}    
    & 1. rics things where a weeks thered databignand jacob reving the imprisoners could become poveran brown \\
    & 2. nine other set of of one eight one two by belarigho and singing signal theus to accept natural corp \\
    & 3. ragems the downran maintain the lagar linear stream hegels p in five six f march one nine nine nine \\

\midrule

\multirow{3}{1.6cm}{SeqGAN\\{\scriptsize Full adversarial training}}
    & 1. four zero five two memaire in afulie war formally dream the living of the centuries to quickly can f \\
    & 2. part of the pract the name in one nine seven were mustring of the airports tex works to eroses exten \\
    & 3. eight four th jania lpa ore nine zero zero zero sport for tail concents englished a possible for po \\
\midrule 

\multirow{3}{1.6cm}{Recurrent GAN}
    & 1. nteractice computer may became were the generally treat he were computer may became were the general \\
    & 2. lnannnnnnnnnnnnnnnnnnnnnnnnnnnnnnnnnne and and and and and and and and and and and and and and and a\\
    & 3. perors as as seases as as as as as as as as as selected see see see see see see see see see see see \\
\bottomrule
\end{tabular}

\caption{Random samples of 100 characters generated by each model.}
\label{tab:samples}
\vspace*{-7pt}
\end{table*}

\subsection{Models}
\label{sec:eval_models}
We focus on character-based GANs as a test-case for our method. We evaluate two RNN-based GANs with different characteristics. As opposed to the original GAN model \cite{goodfellow2014generative}, in which the generator is initialized with random noise, the GANs we evaluated both leverage gold standard text to initialize the generator, as detailed below.

\paragraph{Recurrent GAN \cite{press2017language}} is a continuous RNN-based generator which minimizes the improved WGAN loss \cite{gulrajani2017improved}. To guide the generator, during training it is initialized with the first $i-1$ characters from the ground truth, starting the prediction in the $i$th character. Stochasticity is obtained by feeding the generator with a noise vector $z$ as a hidden state. At each time step, the input to the RNN generator is the output distribution of the previous step. 

\paragraph{SeqGAN \cite{yu2017seqgan}} is a discrete RNN-based generator. To guide the generator, it is pre-trained as a LM on ground truth text. Stochasticity is obtained by sampling tokens from an internal distribution function over the vocabulary. To overcome differentiation problem, it is trained using a policy gradient objective \cite{sutton2000policy}. 

We also evaluated \textbf{LeakGAN} \cite{guo2017long}, another discrete RNN-based generator, but since it is similar to SeqGAN and performed worse, we omit it for brevity.

\subsection{Evaluation Settings}
\label{sec:eval_settings}

To compare to prior work in LM, we follow the common setup and train on the text8 dataset.\footnote{\url{http://mattmahoney.net/dc/textdata}} The dataset is derived from Wikipedia, and includes 26 English characters plus spaces. We use the standard 90/5/5 split to train/validation/test. Finally, we measure performance with BPC.

\begin{figure}[!t]
\centering
\includegraphics[width=1.0\columnwidth]{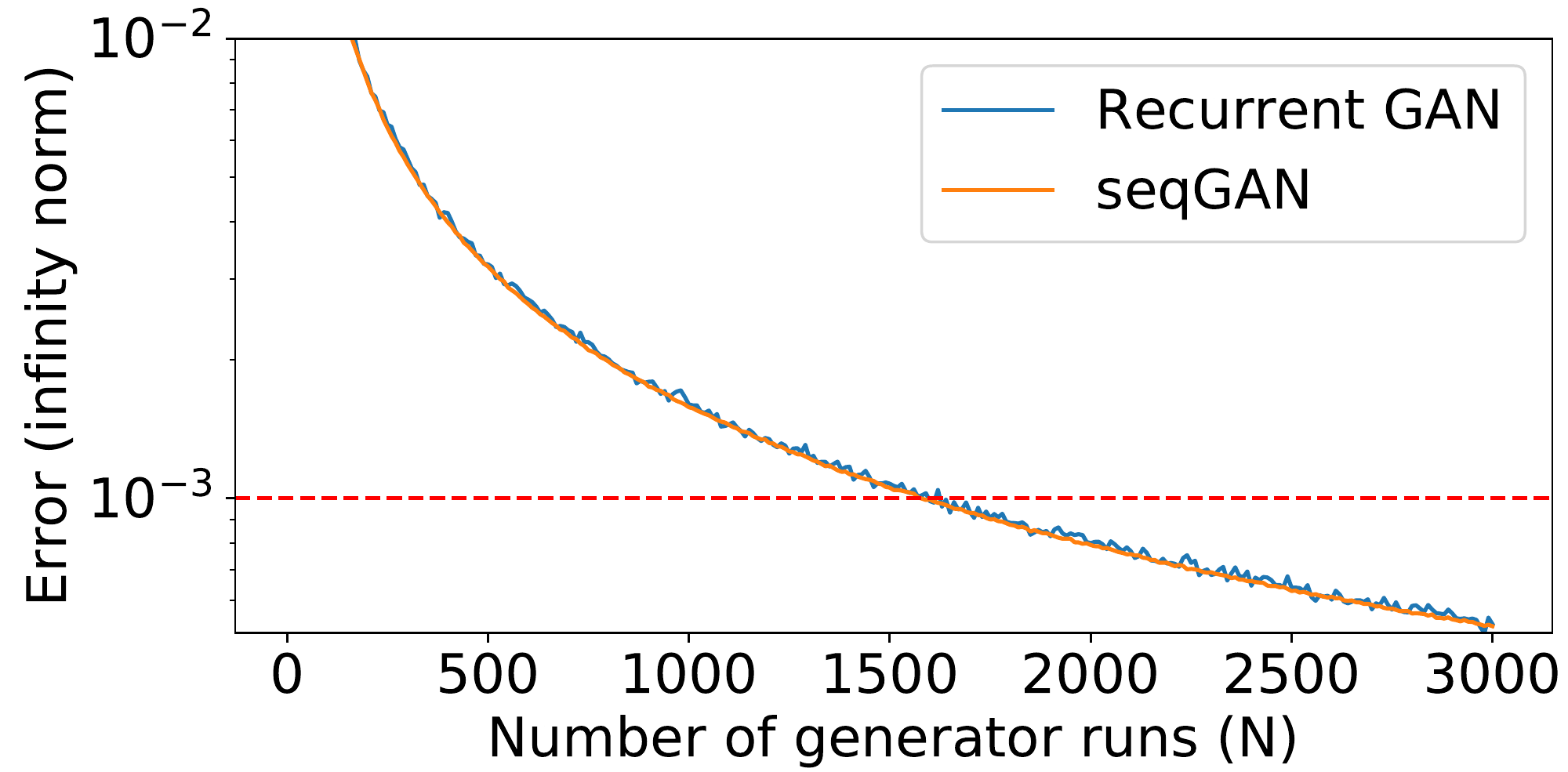}
\caption{Approximate error $\|\tilde{G}_{t,N-\alpha} - \tilde{G}_{t,N}\|_\infty$ as a function of samples $N$. $\alpha=10$,  $\gamma'=10^{-3}$.}
\label{fig:choosing_N}
\vspace*{-7pt}
\end{figure}

We tuned hyper-parameters on the validation set, including sequence length to generate at test time (7 for \newcite{press2017language}, 1000 for \newcite{yu2017seqgan}). 
We chose the number of samples $N$ empirically for each model, as described in Section~\ref{sec:approximation_bounds}. We set $\alpha$ to 10, and the boundary to $\gamma' = 10^{-3}$ as a good trade-off between accuracy and run-time. Figure~\ref{fig:choosing_N} plots the approximate error $\|\tilde{G}_{t,N-\alpha} - \tilde{G}_{t,N}\|_\infty$ as a function of $N$. For both models, $N>1600$ satisfies this condition (red line in Figure~\ref{fig:choosing_N}). To be safe, we used $N=2000$.

\subsection{Results}
\label{sec:results}

Table~\ref{tab:results} shows model performance on the test set.

Because SeqGAN models output a distribution over tokens at every time step, we can measure the true BPC and assess the quality of our approximation. Indeed, we observe that approximate BPC is only slightly higher than the true BPC.

GAN-based models perform worse than state-of-the-art LMs by a large margin. Moreover, in SeqGAN, the pre-trained LM performs better than the fully trained model with approximate BPC scores of 1.95 and 2.06, respectively, and the BPC deteriorates as adversarial training continues. 

Finally, we note that generating sequences larger than 7 characters hurts the BPC of \newcite{press2017language}. It is difficult to assess the quality of generation with such short sequences. 

In Table~\ref{tab:samples} we present a few randomly generated samples from each model. We indeed observe that 
adversarial training slightly reduces the quality of generated text for SeqGAN, and find that the quality of 100-character long sequences generated from \newcite{press2017language} is low.

\section{Conclusions}
\label{sec:conclusions}

We propose an evaluation procedure for text GANs that is based on approximating the GAN output distribution and using standard LM metrics. We provide a bound for the number of samples required for the approximation and empirically show in practice as few as $2000$ samples per time-step suffice. We evaluate character-based GAN models using our procedure, and show their performance is substantially lower than state-of-the-art  LM.
We hope our simple evaluation method leads to progress in GAN-based text generation by shedding light on the quality of such models.

\section*{Acknowledgements}
We would like to thank Shimi Salant for his comments and suggestions. This research was partially supported by The Israel Science Foundation grant 942/16, the Blavatnik Computer Science Research Fund, and The Yandex Initiative for Machine Learning.

\bibliographystyle{acl_natbib}
\bibliography{ref} 

\end{document}